%% file: main.tex
\documentclass[11pt]{article}
\usepackage{emnlp2020}
\usepackage{times}
\usepackage{url}
\usepackage{latexsym}
\usepackage{xspace}

\usepackage{url}
\usepackage{booktabs}       
\usepackage{amsfonts}       
\usepackage{nicefrac}       
\usepackage{microtype}      
\usepackage{amsmath}
\usepackage{color}
\usepackage{multirow}
\usepackage{graphicx}
\usepackage{csquotes}
\usepackage{subfig}
\usepackage{algorithm2e}
\usepackage{enumitem}
\usepackage{gensymb}
\usepackage{natbib}

\renewcommand{\eqref}[1]{Eq.~(\ref{#1})}

\DeclareMathOperator*{\argmin}{argmin}
\DeclareMathOperator*{\argmax}{argmax}

\newcommand{\mage}{\textsc{MLG}\xspace}
\newcommand{\mageoneVII}{\textsc{SLG 7}\xspace}

\newcommand{\mageIII}{\textsc{MLG 5-7}\xspace}
\newcommand{\mageIV}{\textsc{MLG 5-8}\xspace}
\newcommand{\mageIVcoarse}{\textsc{MLG 4-7}\xspace}

\newcommand{\ourcamcoder}{\textsc{CamCoder}\xspace}
\newcommand{\popbase}{\textsc{PopBaseline}\xspace}

\def\pop{\operatorname{pop}}
\def\dist{\operatorname{dist}}

\long\def\commented#1{{}}

\newcommand{\footremember}[2]{%
   \thanks{\xspace\xspace#2}
    \newcounter{#1}
    \setcounter{#1}{\value{footnote}}%
}
\newcommand{\footrecall}[1]{%
    \footnotemark[\value{#1}]%
} 

\aclfinalcopy

\title{Spatial Language Representation with  Multi-Level Geocoding}

\author{
    Sayali Kulkarni\footremember{eq}{Equal contribution} \\
        Google Research \\
        {\tt \small{sayali@google.com}} \\ \And   
    Shailee Jain\footrecall{eq} \footremember{intern}{Work done during internship at Google} \\
        University of Texas, Austin \\
        {\tt \small{shailee@cs.utexas.edu}} \\ \And 
    Mohammad Javad Hosseini\footrecall{intern} \\
        University of Edinburgh \\
        {\tt \small{javad.hosseini@ed.ac.uk}} \\ \AND
    Jason Baldridge \\
        Google Research \\
        {\tt \small{jasonbaldridge@google.com}} \\ \And
    Eugene Ie \\
        Google Research \\
        {\tt \small{eugeneie@google.com}} \\ \And
    Li Zhang \\
        Google Research \\
        {\tt \small{liqzhang@google.com}}
}

\date{}

\begin{document}
\maketitle
\begin{abstract}
We present a multi-level geocoding model (MLG) that learns to associate texts to geographic locations. The Earth's surface is represented using space-filling curves that decompose the sphere into a hierarchy of similarly sized, non-overlapping cells. MLG balances generalization and accuracy by combining losses across multiple levels and predicting cells at each level simultaneously. Without using any dataset-specific tuning, we show that MLG obtains state-of-the-art results for toponym resolution on three English datasets. Furthermore, it obtains large gains without any knowledge base metadata, demonstrating that it can effectively learn the connection between text spans and coordinates---and thus can be extended to toponymns not present in knowledge bases. 
\end{abstract}

\input{01_intro}
\input{02_spatialreps}

\input{03_model}
\input{04_evalsetup}

\input{05_experiments}
\input{06_future}

\bibliography{mage_earthsea}
\bibliographystyle{acl_natbib}

\end{document}

%% file: 01_intro.tex
\section{Introduction}

Geocoding is the task of resolving a location reference in text to a corresponding point or region on Earth. It is often studied in the context of social networks, where metadata and the network itself provide additional signals to geolocate nodes (usually people) \citep{Backstrom:2010:GeolocationFacebook,Rahimi:2015:Geocoding}. These evaluate performance on social media data, which has a strong bias for highly-populated locations. If the locations can be mapped to an entity in a knowledge graph, toponym resolution -- which is a special case of entity resolution -- can be used to resolve location references to geo-coordinates, This can be done using heuristics based on both location popularity \citep{Leidner:2007:ToponymRI} and distance between candidate locations \citep{Speriosu:2013:ToponymResolution}, as well as learning associations from text to locations. 

\begin{figure}
    \centering
    \includegraphics[width=0.48\textwidth,viewport=70 135 770 535,clip=true]{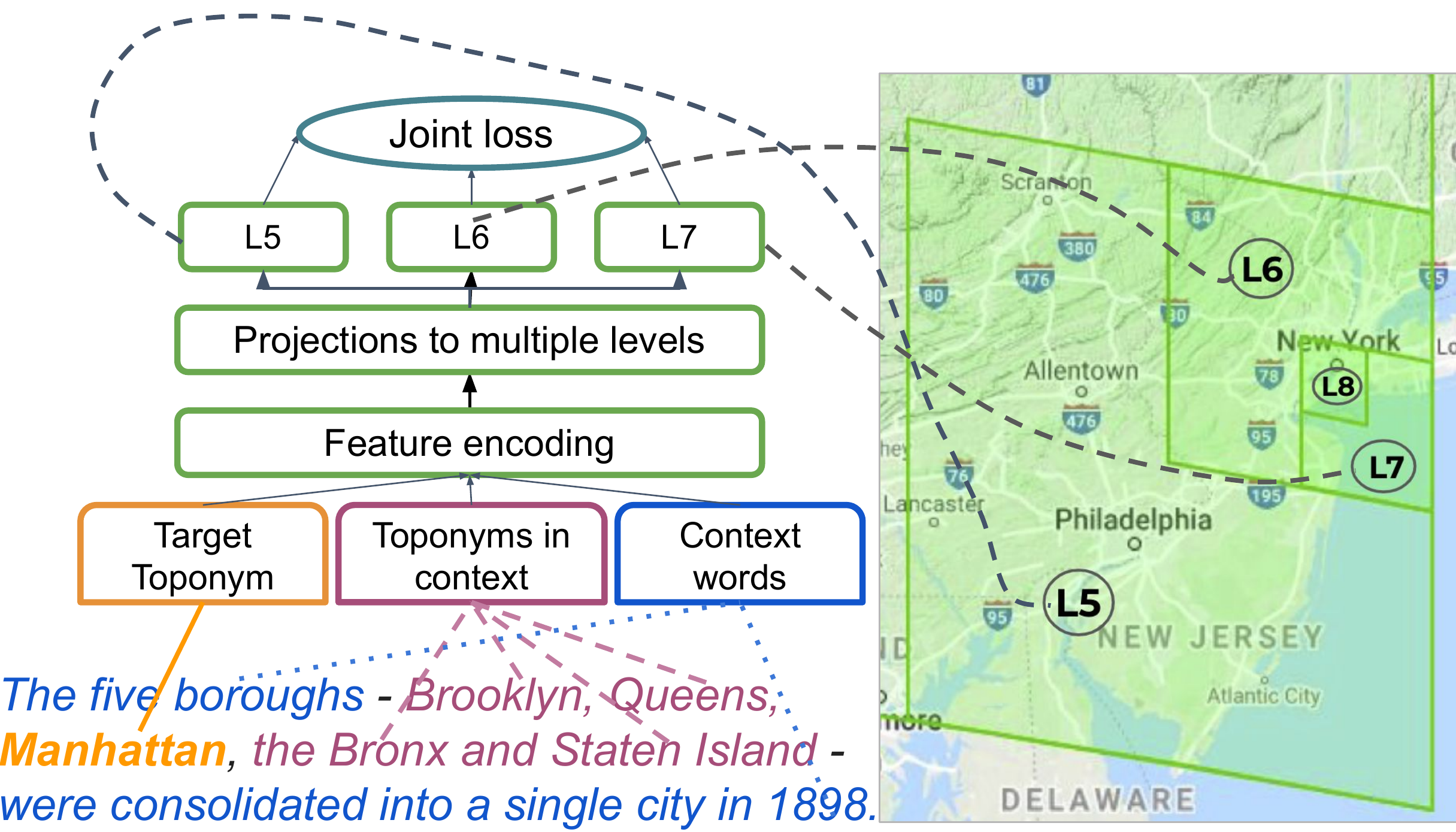}
    \caption{Overview of Multi-Level Geocoder, using multiple context features and jointly predicting cells at multiple levels of the S2 hierarchy.}
    \label{fig:mlg_overview}
\end{figure}

We present Multi-Level Geocoder (MLG, Fig. \ref{fig:mlg_overview}), a model that learns spatial language representations that map toponyms from English texts to coordinates on Earth's surface. This geocoder is not restricted to resolving toponyms to specific location \textit{entities}, but rather to geo-coordinates directly. MLG can thus be extended to non-standard location references in future. For comparative evaluation, we use three toponym resolution datasets from distinct textual domains. MLG shows strong performance, especially when gazetteer metadata and population signals are unavailable. 

MLG is a text-to-location neural geocoder similar to \textit{CamCoder} \citep{Gritta:2018:MapVec}. Locations are represented using S2 geometry\footnote{\url{https://s2geometry.io/}}---a hierarchical discretization of the Earth's surface based on space-filling curves. S2 naturally supports spatial representation at multiple levels, including very fine grained cells (as small as ~1cm$^2$ at level 30); here, we use different combinations of levels 4 ($\sim$300K km$^2$) to 8 ($\sim$1K km$^2$). MLG predicts the classes at multiple S2 levels by jointly optimizing for the loss at each level to balance between generalization and accuracy. 
The example shown in Fig. \ref{fig:mlg_overview} covers an area around New York City by cell id \texttt{0x89c25} at level 8 and \texttt{0x89c4} at level 5. This is more fine-grained than previous work, which tends to use arbitrary square-degree cells, e.g. 2$\degree$-by-2$\degree$ cells (~48K km$^2$) \citep{Gritta:2018:MapVec}. The hierarchical geolocation model over $kd$-trees of \citep{Wing:2014:GelocationHierarchical} can have some more fine-grained cells, but we predict over a much larger set of finer cells. Furthermore, we train a single model that jointly incorporates multi-level predictions rather than learning many independent per-cell models and do not rely on gazetteer-based features.

We consider toponymn resolution for evaluation, but focus on distance-based metrics rather than standard resolution task metrics like top-1 accuracy. When analyzing three common evaluation sets, we found inconsistencies in the true coordinates that we fix and unify to support better evaluation.\footnote{We will release these. Please contact the first author in the meantime if interested.}

%% file: 02_spatialreps.tex
\section{Spatial representations and models}
\label{section:spatial-repn}

Geocoders maps a text span to geo-coordinates---a prediction over a continuous space representing the surface of a (almost) sphere. We adopt the standard approach of quantizing the Earth's surface as a grid and performing multi-class prediction over the grid's cells. There have been studies to model locations as standard bivariate Gaussians on multiple flattened regions \citep{eisenstein-etal-2010-latent,priedhorsky:etal:2014}), but this involves difficult trade-offs between flattened region sizes and the level of distortion they introduce.

We construct a hierarchical grid using the S2 library. S2 projects the six faces of a cube onto the Earth's surface and each face is recursively divided into 4 quadrants, as shown in Figure \ref{fig:mlg_overview}. Cells at each level are indexed using a Hilbert curve. Each S2 cell is represented as a 32-bit unsigned integer and can represent spaces as granular as $\approx 1cm^2$. S2 cells preserves cell size across the globe compared to commonly used degree-square grids (e.g. $1^{\degree}$x$1^{\degree}$) \citep{serdyukov:etal:2009,Wing:2011:GeolocationGrids}. Hierarchical triangular meshes \citep{triangular-mesh} and Hierarchical Equal Area isoLatitude Pixelation \citep{Melo:2015:GeocodingDocumentsHierarhical} are alternatives, though lack the spatial properties of S2 cells. Furthermore, S2 libraries provide excellent tooling.

Adaptive, variable shaped cells based on $k$-d trees \citep{Roller:2012:Geolocation} perform well but depend on the locations of labeled examples in a training resource; as such, a $k$-d tree grid may not generalize well to examples with different distributions from training resources. Spatial hierarchies based on containment relations among entities relies heavily on metadata like GeoNames \citep{Kamalloo2018WWW}. Polygons for geopolitical entities such as city, state, and country \citep{Martins2015ExpandingTU} are perhaps ideal, but these too require detailed metadata for all toponyms, managing non-uniformity of the polygons, and general facility with GIS tools. The Point-to-City (P2C) method applies an iterative $k$-d tree-based method for clustering coordinates and associating them with cities\citep{fornaciari2019identifying}. S2 can represent such hierarchies in various levels without relying on external metadata.

Some of the early models used with grid-based representations were probabilistic language models that produce document likelihoods in different geospatial cells \citep{serdyukov:etal:2009,Wing:2011:GeolocationGrids,Dias:2012:Geocoding,Roller:2012:Geolocation}. Extensions include domain adapting language models from various sources \citep{VanLaere:2014:GeoreferencingWikipedia}, hierarchical discriminative models \citep{Wing:2014:GelocationHierarchical,Melo:2015:GeocodingDocumentsHierarhical}, and smoothing sparse grids with Gaussian priors \citep{Hulden:2015:GeolocationKernelDensity}. Alternatively, \citet{fornaciari-hovy-2019-geolocation} use a multi-task learning setup that assigns probabilities across grids and also predicts the true location through regression. \citet{Melo:2017:GeocodingDocSurvey} covers a broad survey of document geocoding. Much of this work has been conducted on social media data like Twitter, where additional information beyond the text---such as the network connections and user and document metadata---have been used \citep{Backstrom:2010:GeolocationFacebook,Cheng:2010:GeolocationTwitter,Han:2014:GeolocateTwitterUser,Rahimi:2015:Geocoding,Rahimi:2016:pigeo,Rahimi:2017:Geolocation}. MLG is not trained on social media data and hence, does not need additional network information. Further, the data does not have a character limit like tweets, so models can learn from long text sequences.

Toponym resolution identifies place mentions in text and predicting the precise geo-entity in a knowledge base \citep{Leidner:2007:ToponymRI,Gritta:2018:GeoParsingEvalSurvey}. The knowledge base is then used to obtain the geo-coordinates of the predicted entity for the geocoding task. Rule-based toponym resolvers \citep{Smith:2001:Toponym,Grover:2010:GeoreferencingHistory,Tobin:2010:GeorefEval,Karimzadeh:2013:GeoTxt} rely on hand-built heuristics like population from metadata resources like Wikipedia and  GeoNames\footnote{\url{www.geonames.org}} gazetteer. This works well for many common places, but it is brittle and cannot handle unknown or uncommon place names. As such, machine learned approaches that use toponym context features have demonstrated better performance \citep{Speriosu:2013:ToponymResolution,Wei:2014:GeocodingTweets,DeLozier:2015:ToponymWordProfiles,Santos:2015:ToponymResolution}. A straightforward--but data hungry--approach learns a collection of multi-class classifiers, one per toponym with a gazetteer's locations for the toponym as the classes (e.g., the WISTR model of \citet{Speriosu:2013:ToponymResolution}). 

A hybrid approach that combines learning and heuristics by predicting a distribution over the grid cells and then filtering the scores through a gazetteer works for systems like TRIPDL \citep{Speriosu:2013:ToponymResolution} and TopoCluster \citep{DeLozier:2015:ToponymWordProfiles}. A combination of classification and regression loss to predict over recursively partitioned regions shows promising results with in-domain training \citep{Cardoso:2019:RNNsForTopo}. CamCoder \citep{Gritta:2018:MapVec} uses this strategy with a much stronger neural model and achieves state-of-the-art results, including gazetteer signals at training time. 

Our experiments go as far as S2 level eight (of thirty), but our approach is extendable to any level of granularity and could support very fine-grained locations like buildings and landmarks. The built-in hierarchical nature of S2 cells makes it well suited as a scaffold for models that learn and combine evidence from multiple levels. This combines the best of both worlds: specificity at finer levels and aggregation/smoothing at coarser levels. 

\begin{table}
    \centering
    \begin{tabular}{rrr}
S2 Level & number of cells & Avg area \\ 
\hline
L4 & 1.5k & 332 \\
L5 & 6.0k & 83 \\
L6 & 24.0k & 21 \\
L7 & 98.0k & 5 \\
L8 & 393.0k & 1 \\
\end{tabular}
     \caption{S2 levels used in MLG. Average area is in 1k km$^2$.}
    \label{tab:s2cells}
\end{table}

%% file: 03_model.tex
\begin{figure*}
    \centering
    \includegraphics[width=0.85\textwidth, viewport=20 115 850 520, clip=true]{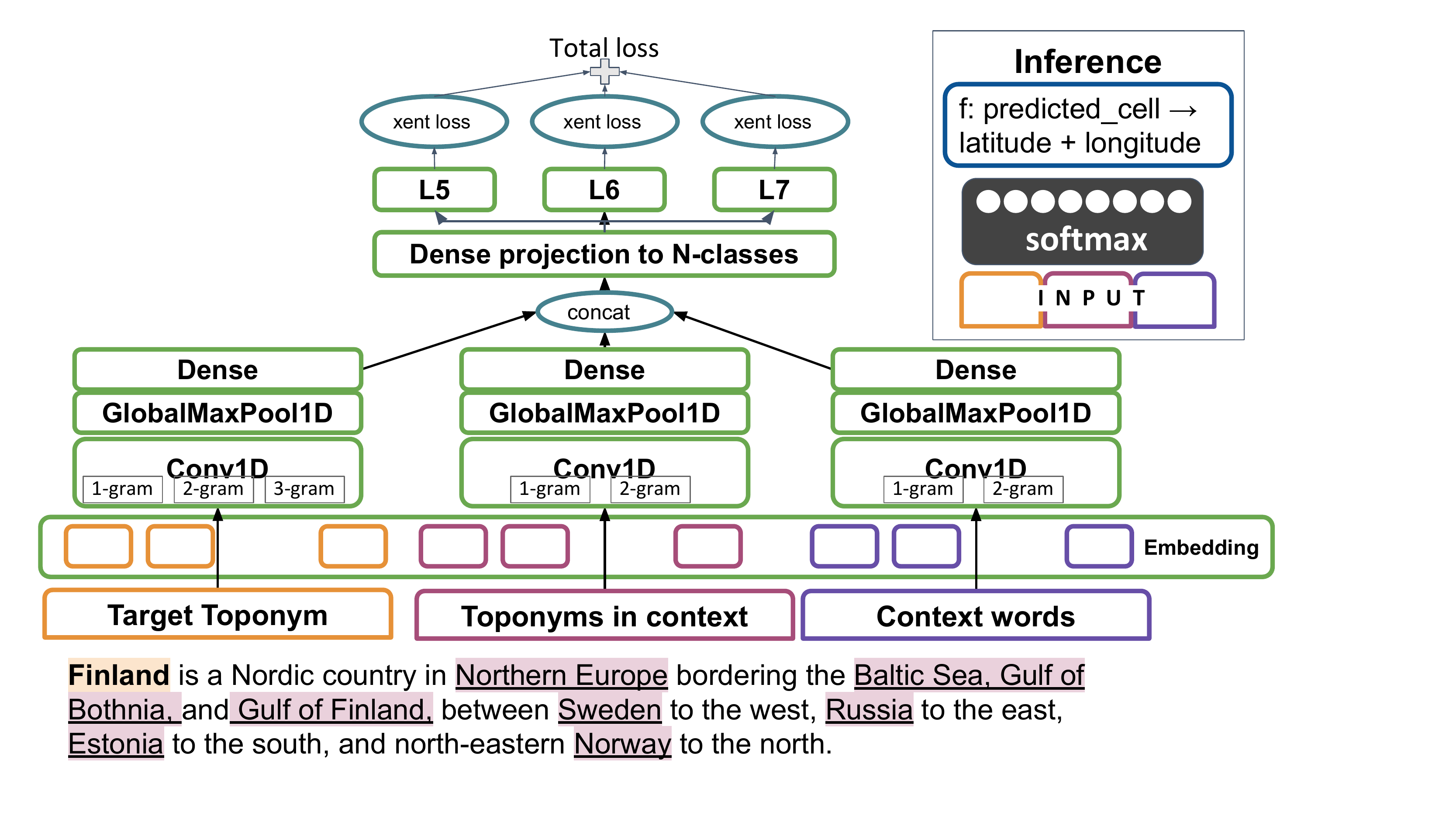}
    \caption{Multi-Level Geocoder model architecture and inference setup.}
    \label{fig:arch}
\end{figure*}

\section{Multi-Level Geocoder (MLG)}
Multi-Level Geocoder (\mage, Figure \ref{fig:arch}) is a text-to-location CNN-based neural geocoder. Context features are similar to CamCoder \citep{Gritta:2018:MapVec} but we do not rely on its metadata-based MapVec feature. The locations are represented using a hierarchical S2 grid that enables us to do multi-level prediction jointly, by optimizing for total loss computed from each level.

\subsection{Building blocks}
For a toponym in context, MLG predicts a distribution over all cells via a convolutional neural network. \textit{Optionally}, the predictions may be snapped to the closest valid cells that overlap the gazetteer locations for the toponym, weighted by their population similar to CamCoder. CamCoder incorporates side metadata in the form of its \textit{MapVec} feature vector, which encodes knowledge of potential locations and their populations matching all toponym in the text. For each toponym, the cells of all candidate locations are activated and given a prior probability proportional to the highest population. These probabilities are summed over all toponyms and renormalized as \textit{MapVec} input. It thus uses population signals in both the MapVec feature in training and in output predictions. This biases predictions toward locations with larger populations---which MLG is not prone to do. 

\subsection{Multi-level classification}

\mage's core is a standard multi-class classifier using a CNN. We use multi-level S2 cells as the output space from a multi-headed model. The penultimate layer maps representations of the input to probabilities over S2 cells. Gradient updates are computed using the cross entropy loss between the predicted probabilities $\mathbf{p}$ and the one-hot true class vector $\mathbf{c}$. 

\mage\ exploits the natural hierarchy of the geographic locations by jointly predicting at different levels of granularity. CamCoder uses 7,823 output classes representing 2x2 degree tiles, after filtering cells that have no support in training, such as over bodies of water. This requires maintaining a cumbersome mapping between actual grid cells and the classes. \mage's multi-level hierarchical representation overcomes this problem by including coarser levels. Here, we focus on three levels of granularity: L5, L6 and L7 (shown in Table \ref{tab:s2cells}), each giving 6K, 24K, and 98K output classes, respectively. 

We define losses at each level (L5, L6, L7) and minimize them jointly, i.e., $\mathcal{L}_\text{total} = (\mathcal{L}(\mathbf{p}_{\text{L5}}, \mathbf{c}_{\text{L5}}) + \mathcal{L}(\mathbf{p}_{\text{L6}}, \mathbf{c}_{\text{L6}}) + \mathcal{L}(\mathbf{p}_{\text{L7}}, \mathbf{c}_{\text{L7}}))/3$. At inference time, a single forward pass computes probabilities at all three levels. The final score for each L7 cell is dependent on its predicted probability as well as the probabilities in its corresponding parent L6 and L5 cells. 
Then the final score for $s_\text{L7}(f) = \mathbf{p}_{\text{L7}}(f) * \mathbf{p}_{\text{L6}}(e) * \mathbf{p}_{\text{L5}}(d)$ and the final prediction is $\hat{y} = \argmax_y s_\text{L7}(y)$. This approach is easily extensible to capture additional levels of resolution---we also present results with finer resolution at L8, with $\sim$1K km$^2$ area and coarser resolution at L4 with $\sim$300K km$^2$ area for comparison. 

\mage\ consumes three features extracted from the context window: (a) token sequence ($w_{a,1:l}$), (b) toponym mentions ($w_{b,1:l}$), and (c) surface form of the target toponym ($w_{c,1:l}$). All text inputs are transformed uniformly, using shared model parameters. Let input text content be denoted as a word sequence $w_{x,1:l}=[w_{x,1}, \dots, w_{x,l}]$, initialized using GloVe embeddings $\phi(w_{x,1:l})=[\phi(w_{x,1}), \dots, \phi(w_{x,l})]$ \cite{Pennington:2014:Glove}. We use 1D convolutional filters to capture n-gram sequences through $\texttt{Conv1D}_n(\cdot)$. This is followed by max pooling which is projected onto a dense layer to get  $\texttt{Dense}(\texttt{MaxPool}(\texttt{Conv1D}_n(\phi(w_{x,1:l})))) \in \mathbb{R}^{2048}$, where $n=\{1, 2\}$ for the sequence of tokens and toponym mentions, and $n=\{1, 2, 3\}$ for the target toponym. These projections are concatenated as input representation.

\subsection{Gazetteer-constrained prediction}
\label{sec:gazfilter}

The only way \mage uses geographic information is from training labels for toponym targets. At test time, \mage\ predicts a distribution over all cells at each S2 level given the input features and picks the highest probability cell at the most granular level. We use the center of the cell as predicted coordinates. However, when the goal is to resolve a specific toponym, an effective heuristic is to use a gazetteer to filter the output predictions to only those that are valid for the toponym. Furthermore, gazetteers come with population information that can be used to nudge predictions toward locations with high populations---which tend to be discussed more than less populous alternatives. Like \newcite{DeLozier:2015:ToponymWordProfiles}, we consider both gazetteer-free and gazetteer-constrained predictions.

Gazetteer-constrained prediction makes toponym resolution a sub-problem of entity resolution. As with broader entity resolution, a strong baseline is an alias table (the gazetteer) with a popularity prior. For geographic data, the population of each location is an effective quantity for characterizing popularity: choosing Paris, France rather than Paris, Texas for the toponym \textit{Paris} is a better bet. This is especially true for zero-shot evaluation where one has no in-domain training data. 

We follow the strategy of \newcite{Gritta:2018:MapVec} for gazetteer constrained predictions. We construct an alias table which maps each mention $m$ to a set of candidate locations, denoted by $C(m)$ using link information from Wikipedia and the population $\pop(\ell)$ for each location $\ell$ is read from WikiData.\footnote{\url{http://www.wikidata.org}} For each of the gazetteer's candidate locations we compute a population discounted distance from the geocoder's predicted location $p$ and choose the one with smaller value as $\argmin_{\ell\in C(m)} \dist(p, \ell) \cdot (1 - c\cdot \pop(\ell) / \pop(m))$. Here, $\pop(m)$ is the maximum population among all candidates for mention $m$, $\dist(p, \ell)$ is the great circle distance between prediction $p$ and location $\ell$, and $c$ is a constant in $[0, 1]$ that indicates the degree of population bias applied. For $c{=}0$, the location nearest the the prediction is chosen (ignoring population); for $c{=}1$, the most populous location is chosen, (ignoring $p$). This is set to 0.90 as found optimal on development set. 

\subsection{Training Data and Representation}
\label{subsec:data-and-repn}

\mage\ is trained on geographically annotated Wikipedia pages, excluding all pages in the WikToR dataset (see sec. \ref{sec:datasets}). For each page with latitude and longitude coordinates, we consider context windows of up to 400 tokens (respecting sentence boundaries) as potential training example candidates. Only context windows that contain the target Wikipedia toponym are used. We use Google Cloud Natural Language API libraries to tokenize\footnote{\url{https://cloud.google.com/natural-language/docs/analyzing-syntax}} the page text and for identifying\footnote{\url{https://cloud.google.com/natural-language/docs/analyzing-entities}} toponyms in the contexts. We use the July 2019 English Wikipedia dump, which has 1.11M location annotated pages giving 1.76M training examples. This is split 90/10 for training/development.

As an example, consider a short context for \textit{United Kingdom}, ``\textit{The UK consists of four constituent countries: England, Scotland, Wales and Northern Ireland.}''. Tokens in the context are lower cased and used as features, e.g., [``\textit{the}'', ``\textit{uk}'', ``\textit{consists}', ..., ``.'']. Sub-strings referring to locations are recognized, extracted and used as features, e.g., [``\textit{uk}'', ``\textit{england}'', ``\textit{scotland}'', ..., ``\textit{ireland}'']. Finally, the surface form of the target mention ``\textit{uk}'' is used as the third feature. 

%% file: 04_evalsetup.tex
\section{Evaluation}

We train \mage\ as a general purpose geocoder and evaluate it on toponym resolution. A strong baseline is to choose the most populous candidate location (\popbase): i.e. $\argmax_{\ell\in C(m)} \pop(\ell)$
\subsection{Evaluation Datasets}
\label{sec:datasets}

We use three public datasets: Wikipedia Toponym Retrieval (WikToR) \cite{Gritta:2018:GeoParsingEvalSurvey}, Local-Global Lexicon (LGL) \cite{Lieberman:2010:LGL}, and GeoVirus \cite{Gritta:2018:MapVec}.  See \newcite{Gritta:2018:GeoParsingEvalSurvey} for extensive discussion of other datasets.

\noindent \textbf{WikToR} (WTR) is the largest programmatically created corpus to date that allows for comprehensive evaluation of toponym resolvers. By construction, ambiguous location mentions were prioritized (e.g. ``\textit{Lima, Peru}'' vs. ``\textit{Lima, Ohio}'' vs. ``\textit{Lima, Oklahoma}'' vs ``\textit{Lima, New York}''). As such, population-based heuristics are counter-productive in WikToR. 

\noindent \textbf{LGL} consists of 588 news articles from 78 different news sources. 
This dataset contains 5,088 toponyms and 41\% of these refer to locations with small populations. About 16\% of the toponyms are for street names, which do not have coordinates; we dropped these from our evaluation set. About 2\% have an entity that does not exist in Wikipedia, which were also dropped. In total, this dataset provides 4,172 examples for evaluation.

\noindent \textbf{GeoVirus} dataset~\cite{Gritta:2018:MapVec} is based on 229 articles from WikiNews.\footnote{\url{https://en.wikinews.org}} The articles detail global disease outbreaks and epidemics and were obtained using keywords such as ``Bird Flu'' and ``Ebola''. Place mentions are manually tagged and assigned Wikipedia page URLs along with their global coordinates. In total, this dataset provides 2,167 toponyms for evaluation. 

WikToR serves as in-domain Wikipedia-based evaluation data, while both LGL and GeoVirus provide out-of-domain news corpora evaluation.

\begin{table*}
\small
    \centering
    \begin{tabular}{ll|rrr|r||rrr|r||rrr|r}
Gaz & & \multicolumn{4}{c||}{AUC of error curve} & 
\multicolumn{4}{c||}{accuracy@161}  & \multicolumn{4}{c}{Mean error}  \\ 
Used & Model & WTR & LGL & GeoV & Avg & WTR & LGL & GeoV & Avg & WTR & LGL & GeoV & Avg \\
\hline
 \multirow{4}{*}{Yes} & \popbase & 66 & 42 & 41 & 50 & 22 & 57 & 68 & 49 & 4175 & 1933 & 898 & 2335 \\ 
 &  \ourcamcoder & 24 & 32 & 15 & 24 & 72 & 63 & 82 & 72 & 440 & 877 & 315 & 544 \\ 
& \mageoneVII & 17 & 28 & \textbf{13} & 19 & 82 & 72 & \textbf{86} & 80 & 480 & 648 & 305 & 478 \\ 
 & \mageIII & \textbf{15} & \textbf{27} & \textbf{13} & \textbf{18} & \textbf{85} & \textbf{73} & 85 & \textbf{81} & \textbf{347} & \textbf{620} & \textbf{276} & \textbf{414} \\ 
\hline
 \multirow{3}{*}{No} & \ourcamcoder & 49 & 60 & 65 & 58 & 70 & 38 & 26 & 45 & 239 & 1419 & 2246 & 1301 \\ 
 & \mageoneVII & 39 & 55 & 56 & 50 & 86 & 49 & 48 & 61 & 424 & 1688 & 1956 & 1356 \\ 
 & \mageIII & \textbf{37} & \textbf{54} & \textbf{55} & \textbf{49} & \textbf{91} & \textbf{53} & \textbf{49} & \textbf{64} & \textbf{180} & \textbf{1407} & \textbf{1690} & \textbf{1092}

    \end{tabular}
    \caption{Comparing population baseline, CamCoder benchmark (our implementation), and our SLG and \mage models on the \textit{unified} data, both with and without the gazetteer filter.}
    \label{tab:base}
\end{table*}

\subsection{Unifying evaluation sets}
We use the publicly available version for the three datasets used in CamCoder. \footnote{\url{https://github.com/milangritta/Geocoding-with-Map-Vector/tree/master/data}} However, after analyzing examples across the evaluation datasets, we identified some inconsistencies in location target coordinates.

First, the WikToR evaluation set delivers annotations given its reliance on GeoNames DB and Wikipedia APIs. However, we discovered that WikToR was mapped from an older version of GeoNames DB which has a known issue of sign flip in either latitude or longitude of some locations. As an example, \textit{Santa Cruz, New Mexico} is incorrectly tagged as (35, 106) instead of (35, -106). This affects 296 out of 5,000 locations in WikToR---mostly cities in the United States and a few in Australia.

\begin{table*}
    \centering
    \small
    \begin{tabular}{l|rrr|r||rrr|r||rrr|r}
 & \multicolumn{4}{c||}{AUC of error curve} & 
\multicolumn{4}{c||}{accuracy@161}  & \multicolumn{4}{c}{Mean error}  \\ 
Inference & WTR & LGL & GeoV & Avg & WTR & LGL & GeoV & Avg & WTR & LGL & GeoV & Avg \\
\hline
L5-7 & \textbf{37} & \textbf{54} & \textbf{55} & \textbf{49} & \textbf{91} & \textbf{53} & \textbf{49} & \textbf{64} & \textbf{180} & \textbf{1407} & \textbf{1690} & \textbf{1092} \\
Only L5 & 48 & 60 & 62 & 57 & 79 & 45 & 39 & 54 & 285 & 1599 & 1957 & 1280 \\ 
 Only L6 & 43 & 57 & 60 & 53 & 90 & 51 & 44 & 62 & 265 & 1534 & 2003 & 1267 \\ 
 Only L7 & 38 & \textbf{54} & 56 & 50 & 89 & 51 & 48 & 63 & 349 & 1525 & 2014 & 1296  
     \end{tabular}
         \caption{Prediction granularity: performance of \mage\ trained with multi-level loss on L5, L6 and L7 but using single level at inference time.}
    \label{tab:granu}
\end{table*}

Second, there are differences in location target coordinates across the 3 datasets since each of them may have been created differently. For example, Canada is represented as (60.0, -95.0) in GeoVirus, (60.0, -96.0) in LGL and (45.4, -75.7) in WikToR. Since we represent locations as points rather than regions, we choose and apply consistent coordinates for each location across the evaluation sets. For this, we re-annotated all three datasets to unify the coordinates for target toponyms. The annotation was done using the coordinates from Wikidata to be consistent with the training labels.

\begin{table*}
\small
    \centering
    \begin{tabular}{l|r|rrr|r||rrr|r||rrr|r}
 & Dev & \multicolumn{4}{c||}{AUC of error curve} & \multicolumn{4}{c||}{accuracy@161}  & \multicolumn{4}{c}{Mean error}  \\ 
Model & loss & WTR & LGL & GeoV & Avg & WTR & LGL & GeoV & Avg & WTR & LGL & GeoV & Avg \\
\hline
\mageIVcoarse & 8.71 & 37 & 55 & 54 & 49 & 91 & 51 & 51 & 64 & 197 & 1529 & 1570 & 1099 \\ 
\textbf{\mageIII} & \textbf{7.25} & 37 & 54 & 55 & 49 & 91 & 53 & 49 & 64 & 180 & 1407 & 1690 & 1092 \\
\mageIV & 13.28 & 38 & 58 & 67 & 54 & 89 & 45 & 24 & 53 & 272 & 1866 & 3058 & 1732 \\ 
    \end{tabular}
     \caption{Models trained with different granularities help trade-off between accuracy and generalization. Selected model \mageIII is based on optimal performance of the holdout.} 
    \label{tab:level_choices}
\end{table*}

\subsection{Evaluation Metrics}
\label{subsec:metrics}

We use three standard metrics in geocoding: accuracy(or accuracy@161km), mean distance error, and AUC for the error curve. Accuracy is the percentage of toponyms that are resolved to with 161km (100 miles) of their true location. Mean distance error is the average of the distance between the predicted location (center of the predicted S2 cell) and true location of the target toponym. AUC is the area under the discrete curve of sorted log-error distances in the evaluation set. AUC\footnote{Unlike the standard AUC, lower is better for AUC since this is based on the curve of error distances.} is an important metric as it captures the entire distribution of errors and is not sensitive to outliers. It also uses the log of the error distances, which appropriately focuses the metric on smaller error distances for comparing models. 

 \nocite{Jurgens2015AUC}

In this paper, we study the benefits of resolving the toponym over multiple levels of granularity to account for the range of populations, resolution ambiguity, topological shapes and sizes of different toponyms. We leave the shaping of the output class space as future work (e.g., using geopolitical polygons instead of points).

%% file: 05_experiments.tex
\begin{table*}
\small
    \centering
    \begin{tabular}{l|rrr|r||rrr|r||rrr|r}
 & \multicolumn{4}{c||}{AUC of error curve} & \multicolumn{4}{c||}{accuracy@161}  & \multicolumn{4}{c}{Mean error}  \\ 
Ablation & WTR & LGL & GeoV & Avg & WTR & LGL & GeoV & Avg & WTR & LGL & GeoV & Avg \\
\hline
 all features & 37 & 54 & 55 & 49 & 91 & 53 & 49 & 64 & 180 & 1407 & 1690 & 1092 \\ 
 $~~~$ - target & 38 & 60 & 69 & 55 & 91 & 39 & 18 & 49 & 174 & 2032 & 2811 & 1672 \\ 
 $~~~$ - all toponyms & 69 & 75 & 82 & 76 & 29 & 14 & 04 & 16 & 4487 & 4442 & 6360 & 5096 \\
 
     \end{tabular}
     \caption{Effect of ablating location features from the input to demonstrate their importance in \mageIII.}
    \label{tab:ablations}
\end{table*}

\section{Experiments}
\label{sec:results}

\subsection{Training}

MLG is trained using TensorFlow \cite{Abadi:2016:TensorFlow} distributed across 13 P100 GPUs. Each training batch processes 512 examples. The model trains up to 1M steps, although they converge around 500K steps. We found an optimal initial learning rate of $10^{-4}$ decaying exponentially over batches after initial warm-up. For optimization, we use Adam \cite{Kingma:2015:Adam} for stability.

We considered S2 levels 4 through 8, including single level (SLG) and multi-level (MLG) variations. \mage's architecture offers the flexibility of doing multi-level training but performing prediction with just one level. Based on the loss on Wikipedia development split, we chose multi-level training and prediction with levels 5, 6 and 7.

Our focus is geocoding without any gazetteer information at inference time. However, we also show that additional gains can be achieved using gazetteers to select relevant cells for a given toponym, and scale the output using the population bias ($c$) described in section \ref{sec:gazfilter}. 

\subsection{Results}

Table~\ref{tab:base} shows results for the \popbase, \ourcamcoder, SLG and \mage models on all three datasets for all metrics. For \ourcamcoder, SLG and \mage, we include results with and without the use of gazetteer based population bias (sect. \ref{sec:gazfilter}). 

Our results are reported on the unified datasets. The \ourcamcoder results are based on our own implementation and trained on the same examples as \mage training set. 


\paragraph{Overall trends} The most striking result is how well \mage compares to \ourcamcoder without the use of gazetteer, especially on WikToR, a dataset which was specifically designed to counteract population priors. The architecture that \mage inherits from \ourcamcoder is effective for text geocoding, but \mage generalizes better by leaving out the non-lexical MapVec feature and thereby avoiding the influence of the population bias for the toponyms in the context.

\paragraph{Gazetteer-free performance} \mage's fine-grained multi-level learning and prediction pays off across all datasets, both with and without the use of gazetteer. This is particularly pronounced with AUC, where \mage is 6\% better than \ourcamcoder with the filter on an average across the 3 datasets. Without the gazetteer, \mage has an even larger gap of 9\%. It is also clear that MLG is superior to SLG, validating the use of multi-level learning and prediction. 

\begin{table*}
\small
    \centering
    \begin{tabular}{p{7.2cm} p{7.2cm} p{0.2cm}}
    Arlington is a former manor, village and civil parish in the North Devon district of Devon in England. The parish includes the villages of Arlington and Arlington Beccott. ... & Arlington is a city in Gilliam County, Oregon, United States. The account of how the city received its name varies; one tradition claims it was named after the lawyer Nathan Arlington Cornish, ... & \multirow{4}{*}{\includegraphics[height=0.43\textheight, viewport=660 0 710 610, clip=true]{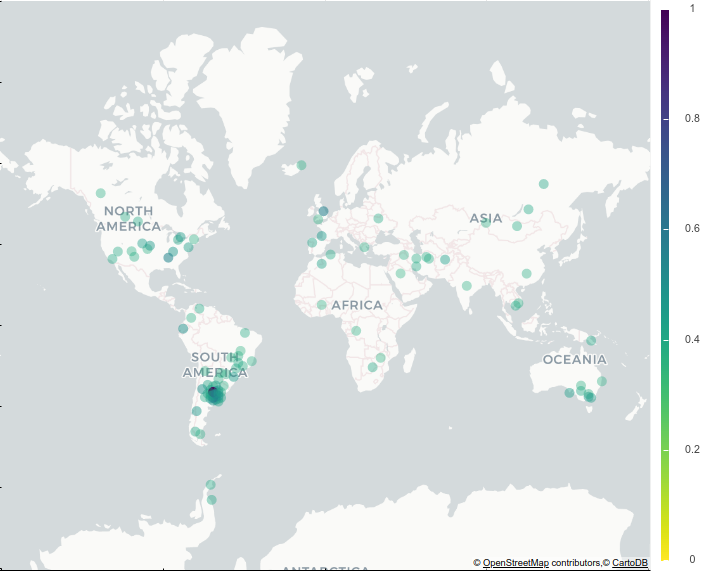}} \\
    
     \includegraphics[width=0.46\textwidth, viewport=0 100 650 450, clip=true]{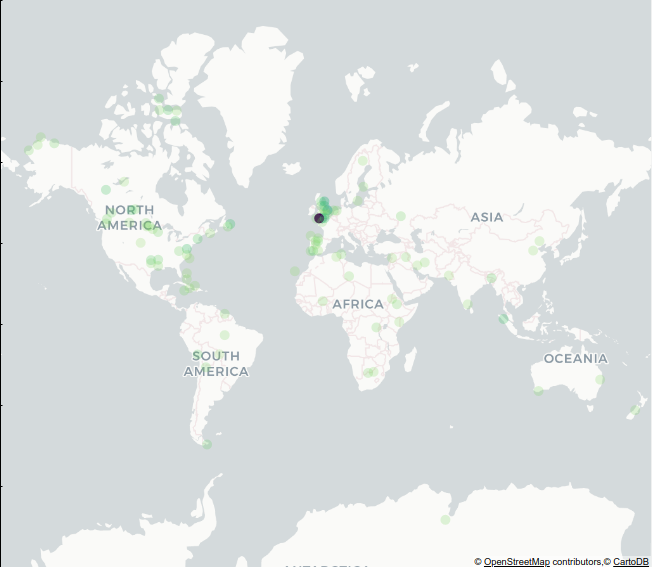} & \includegraphics[width=0.46\textwidth, viewport=0 100 650 450, clip=true]{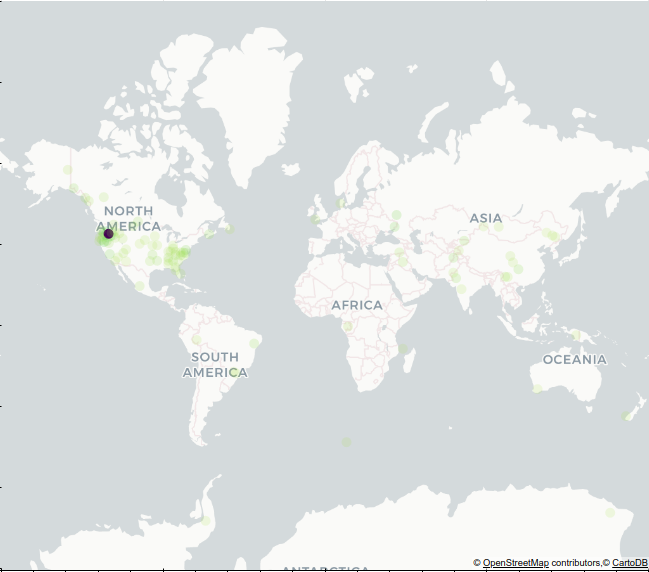} & \\
     \cline{1-2}
     
     Lincoln is a city in Logan County, Illinois, United States. It is the only town in the United States that was named for Abraham Lincoln before he became president.... & Lincoln is a city in the province of Buenos Aires in Argentina. It is the capital of the district of Lincoln (Lincoln Partido). The district of Lincoln was established on ... \\
    \includegraphics[width=0.46\textwidth, viewport=0 100 650 450, clip=true]{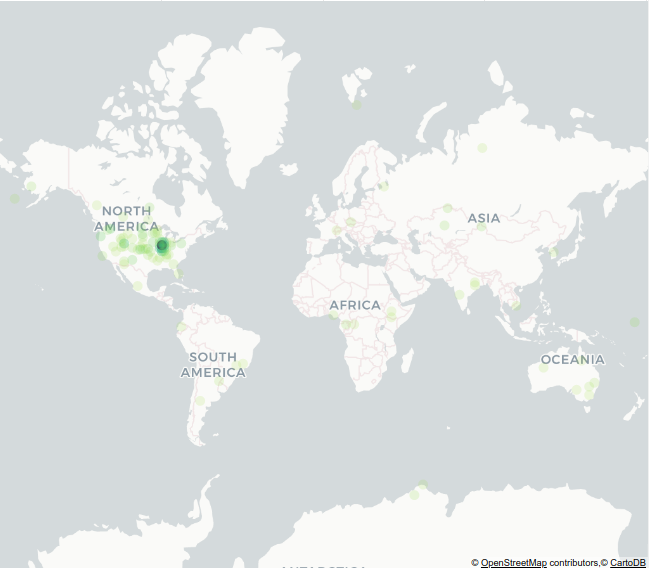} &  \includegraphics[width=0.46\textwidth, viewport=0 100 660 450, clip=true]{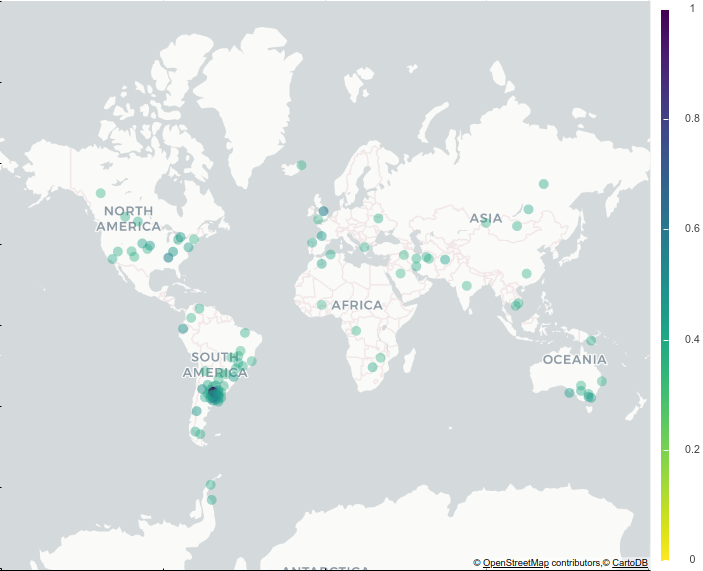} \\
     \end{tabular}
     \caption{Examples showing effect of context on predicted distributions.}
    \label{tab:arlington}
\end{table*}

\paragraph{Model generalization} When not using a gazetteer, \mage is much closer to the strong population baselines for LGL and GeoVirus, indicating that the multi-level approach allows the use of training evidence to generalize better over examples drawn globally (entire world in GeoVirus) as well as locally (the United States of America in LGL).

\paragraph{Multi-level prediction helps.} Table~\ref{tab:granu} compares performance of using individual levels from the same \mage model trained on levels L5, L6 and L7 (and using the gazetteer filter). The tradeoff of predicting at different granularity is clear: when we use lower granularity, e.g. L5 cells, our model can generalize better, but it may be less precise given the large size of the cells. On the other hand, when using finer granularity, e.g. L7 cells, the model can be more accurate in dense regions, but could suffer in sparse regions where there is less training data. Combining the predictions from all levels balances the strengths effectively.

\paragraph{Levels five through seven offer best tradeoff} Table \ref{tab:level_choices} shows performance of \mage\ by training and predicting with multiple levels at different granularities. Overall, using levels five through seven (which has the best development split loss) provides the strongest balance between generalization and specificity. 
For locating cities, states and countries, especially when choosing from candidate locations in a gazetteer, L8 cells do not provide much greater precision than L7 and suffer from fewer examples as evidence in each cell.

\paragraph{Qualitative examples} An effective use of context  in correctly predicting the coordinates is shown in Table \ref{tab:arlington} on two examples - one for `Arlington' and one for `Lincoln'. In both pairs, the context helps to shift the predictions in the right regions on the map. It is not biased by just the most populous place. Here we only show a part of the context for clarity though the actual context longer as described in Section \ref{subsec:data-and-repn}.

\paragraph{Ablations} Table \ref{tab:ablations} provides results when ablating salient features at inference time, removing either the target toponym or all toponyms. While masking the target toponym does not change results much except for GeoVirus, masking all other toponyms does degrade performance considerably. Nevertheless, it may still be possible with just the context words, which include other named entities, characteristics of the place, and location-focused words in few cases. For example, `Arlington (England)', can still be geolocated after all toponyms are masked (Figure \ref{fig:ablated_arlington}); however, the distribution is much more spread out in this case.

%% file: 06_future.tex
\begin{figure}
\centering
    \includegraphics[width=0.46\textwidth, viewport=0 100 650 450, clip=true]{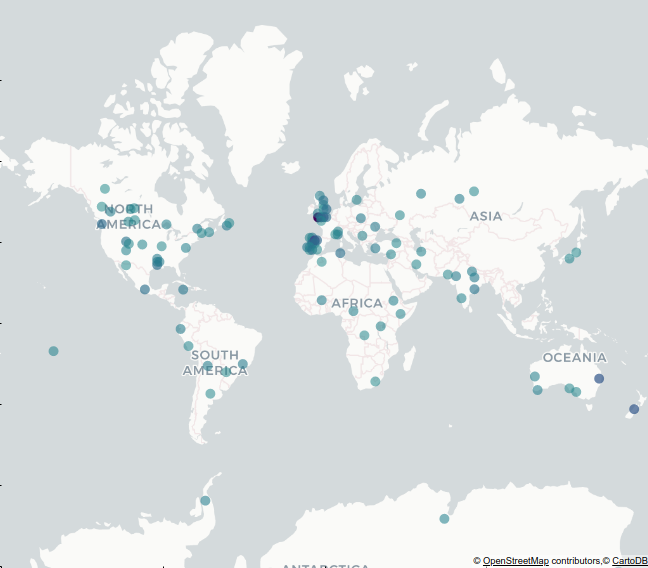}
    \caption{Ablation of all locations from context still leaves other references from context to enable correct prediction.}
    \label{fig:ablated_arlington}
\end{figure}

\section{Future work}
\mage\ uses of multi-level optimization for the inherently hierarchical problem of geocoding for toponym resolution. With just textual feature inputs, we can predict the location of a target toponym--with minimal to no metadata from external gazetteer for inference--with good accuracy. This makes it possible to use \mage\ for corpora where gazetteer information is not available, such as historical texts \cite{delozier-etal-2016-creating}. Further, since the models generalize very well across domains, they can be used in real-time applications like news feeds. While we use the multi-level loss in the objective function, this can be further refined by using approaches like hierarchical softmax~\cite{Morin:2005:HSoftmax} that can replace multiple softmax layers with hierarchical layers to incorporate the conditional probabilities across layers. Another future direction involves smoothing the label space during training to capture the relations among cells that are near one another. This would also enable shaping the output class space to polygons instead of points, which is more realistic for geographical regions. 





